\documentclass{article}
\usepackage[utf8]{inputenc}
\usepackage{amssymb,amsmath}
\usepackage[numbers]{natbib}
\usepackage{graphicx}
\graphicspath{images/}
\usepackage{comment}
\usepackage{hyperref}
\usepackage{xcolor}
\usepackage{authblk}

\title{Capturing Local Temperature Evolution during Additive Manufacturing through Fourier Neural Operators}

\author[1]{Jiangce Chen}
\author[1]{Wenzhuo Xu}
\author[1]{Martha Baldwin}
\author[2]{Bj\"orn Nijhuis}
\author[2]{Ton van den Boogaard}
\author[1]{Noelia Grande Guti\'{e}rrez}
\author[1]{Sneha Prabha Narra}
\author[1]{Christopher McComb\thanks{ccm@cmu.edu \vspace{-1em}Address all correspondence to this author.}}
\affil[1]{Carnegie Mellon University\\
	Pittsburgh, PA, USA}
\affil[2]{University of Twente\\
	Enschede, the Netherlands}

\begin{document}
\maketitle 
\begin{abstract}
High-fidelity, data-driven models that can quickly simulate thermal behavior during additive manufacturing (AM) are crucial for improving the performance of AM technologies in multiple areas, such as part design, process planning, monitoring, and control. However, the complexities of part geometries make it challenging for current models to maintain high accuracy across a wide range of geometries. 
Additionally, many models report a low mean square error (MSE) across the entire domain (part). However, in each time step, most areas of the domain do not experience significant changes in temperature, except for the heat-affected zones near recent depositions. Therefore, the MSE-based fidelity measurement of the models may be overestimated.

This paper presents a data-driven model that uses Fourier Neural Operator to capture the local temperature evolution during the additive manufacturing process. In addition, the authors propose to evaluate the model using the $R^2$ metric, which provides a relative measure of the model's performance compared to using mean temperature as a prediction.
The model was tested on numerical simulations based on the Discontinuous Galerkin Finite Element Method for the Direct Energy Deposition process, and the results demonstrate that the model achieves high fidelity as measured by $R^2$ and maintains generalizability to geometries that were not included in the training process.

\end{abstract}

\section{Introduction}

\begin{figure*}[ht]
    \centering
    \includegraphics[width=\linewidth]{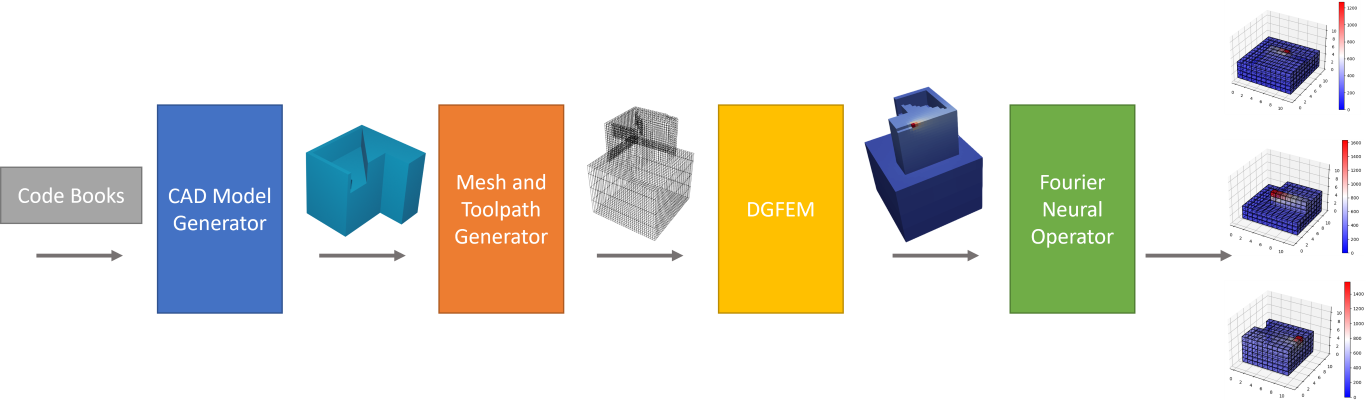}
    \caption{The overview of the framework proposed in this work, which trains a data-driven geometry-agnostic model to capture the local temperature evolution during AM process. An autoregressive generative model, SkexGen \cite{xu2022skexgen},  takes disentangled codebooks to construct CAD models in the variations of encoded topological, geometric, and extrusion features. The hexahedron mesh and the toolpath are generated automaically by a code we developed. A physics-based simulation model based on Discontinuous Galerkin Finite Element Method is employed to generate the ground truths of temperature evolutions. Then, a machine learning model based on Fourier Neural Network is trained to predict the local temperature evolution in the windows around concerned regions. }
    \label{fig:method_overview}
\end{figure*}

Additive manufacturing (AM) has become more than a niche laboratory technology and is becoming increasing vital across various industries. This is largely due to its enormous potential in fabricating complex parts at a lower cost compared to traditional methods like machining or casting. Among the various AM technologies, Directed Energy Deposition (DED) processes have garnered considerable attention because they can build large-scale metal parts up to several meters in size at high deposition rates \cite{lehmann2022large}. This paper therefore focuses on DED processes.

However, the adoption of AM is hindered by concerns regarding product quality and process efficiency \cite{mozaffar2021geometry}. For instance, Glerum et al. \cite{glerum2021mechanical} identified inconsistencies in the AM process where the same process parameters can lead to varying material properties. Thus, establishing process-structure-property relationships for AM has become a critical research area in AM technologies \cite{kouraytem2021modeling, luo2023dataset, popova2017process, gordon2020defect, fang2022data, thanki2022melt}.
The temperature history resulting from the process parameters is a key determinant in several aspects, including melt pool characteristics \cite{cook2020simulation}, formation of defects such as lack of fusion and hot cracking \cite{wang2020evaporation, mukherjee2018mitigation}, metal grain structure \cite{zinoviev2016evolution}, and residual stresses caused by high thermal gradients \cite{dong2021new}. However, obtaining experimental data on temperature history is expensive and also challenging to capture at every point of a part. Therefore, numerical simulation is an attractive, less-expensive alternative for obtaining comprehensive temperature history data at scale.
While many high-fidelity, physics-based simulation models have been developed, such as the method based on Discontinuous Galerkin Finite Element Method (DGFEM) \cite{nijhuis2021efficient}, their time cost is still too high for many crucial applications. For example, iterative optimizations such as design for AM \cite{mozaffar2023differentiable} and process parameter planning \cite{sun2022thermodynamics} require a massive number of thermal simulations. Additionally, real-time simulation is needed for in-situ process monitoring and control \cite{mahmoud2021applications, abouelnour2022situ}.

Therefore, data-driven models have gained attention as a means to quickly simulate thermal behavior during AM. Mozaffar et al. \cite{mozaffar2018data} developed a data-driven machine learning (ML) model based on recurrent neural networks (RNN) to predict the thermal histories of arbitrary points in a part built by the DED process. For real-time temperature prediction, Paul et al. \cite{paul2019real} proposed a framework based on extremely randomized trees (ERT), which is an ensemble of bagged decision trees that use temperatures of prior voxels and laser information as inputs to predict temperatures of subsequent voxels. Stathatos et al. \cite{stathatos2019real} proposed an artificial neural network (ANN) that predicts in real-time the evolution of temperature and density for arbitrary long tracks. Roy et al. \cite{roy2020data} observed that the AM process exhibits a high level of redundancy and periodicity and introduced a geometry representation that extracts features directly from the GCode for a ML model, such as local distances from heat sources and cooling surfaces. However, these models can only achieve good accuracy in predicting the thermal evolution of samples with geometries similar to those in the training dataset and may not perform well on more complex geometries unseen in the training process.

In order to improve the generalizability of ML models for thermal history prediction in complex geometries, Ness et al. \cite{ness2022towards} developed an ERT model that utilizes engineered features based on the underlying physics of the thermal process. Additionally, Mozaffar (2021) \cite{mozaffar2021geometry} proposed a geometry-agnostic data-driven model using (Graph Neural Network) GNNs to capture spatiotemporal dependencies of thermal responses in AM processes. However, current ML models are typically mesh-dependent, which can hinder their adoption in applications with finer meshes than those used for training. For example, models trained on low resolution meshes usually cannot be used to predict temperatures in the high-resolution meshes necessary for in-situ process control. Furthermore, although these models achieve low normalized mean squared error (MSE) over the built part, they may fail to accurately capture the dramatic temperature changes in the heat-affected zones (HAZ) of recent depositions. For instance, experiments in \cite{mozaffar2021geometry} have shown that the difference between the predicted temperature and ground truth in HAZ can be as large as $300^{\circ} C$. Given the close relationship between HAZ temperature and melt-pool dimensions, defect formation, and microstructure formation, accurate HAZ temperature prediction is critical. However, commonly used metrics, such as MSE, may not accurately reflect a model's performance in predicting temperatures in these critical areas, as areas far from the heat source often have stable low temperatures without significant changes. Therefore, even if an ML model predicts low accuracy at the HAZ of recent depositions, it may still score well in MSE-based fidelity measurements.

This paper proposes a data-driven model that utilizes a Fourier Neural Operator to capture the local temperature evolution during additive manufacturing (AM). In addition to the current temperature, the model incorporates the heat source locations and local distances to the cooling surfaces to predict the temperature in the next time step. The contributions of this paper are summarized as follows:
\begin{enumerate}
    \item A mesh-independent ML framework is established for temperature prediction in the AM process.
    \item The ML model prioritizes capturing the temperature evolutions in the Heat Affected Zone (HAZ) near the recent deposition.
    \item An automatic pipeline is build to generate a dataset of geometric models using an autoregressive generative model with customized and existing tools, which are then meshed and have toolpaths generated using code developed by the authors.
\end{enumerate}

A high-fidelity thermal simulation model based on Discontinuous Galerkin Finite Element Method is then applied to obtain temperature histories for use in ML training and testing. The physical coefficients and parameters used in the numerical method have been calibrated with experimental data, as demonstrated in \cite{nijhuis2021efficient}.
In addition,  an R-squared ($R^2$) metric is proposed to measure the performance of ML models for temperature prediction in the AM process, as it provides a relative measure of the model's performance compared to using the mean temperature as a prediction.  Results from numerical experiments demonstrate that the proposed model achieves high fidelity as measured by $R^2$ and maintains generalizability to geometries that were not included in the training process. The model and relevant dataset generation methods developed in this paper have the potential to be implemented in the creation of practical thermal simulation software for industrial applications.

Figure \ref{fig:method_overview} provides an overview of the proposed framework. The remainder of the paper is organized as follows. Section \ref{sec:FNO} briefly introduces the concept of Fourier Neural Operator (FNO) and its application in approximating the solution of partial differential equations (PDEs). In Section \ref{sec:method}, we detail the architecture of our ML model, the loss function and evaluation metrics, and provide a description of the heat-affected windows. Section \ref{sec:data_generation} explains the data generation and preprocessing process for training the ML model and outlines the experiment settings. We present and discuss the experimental results in Section \ref{sec:results}, and conclude in Section \ref{sec:conclude}.

\section{The Background of Fourier Neural Operator}\label{sec:FNO}
A nonlinear operator is defined to be a mapping from a space of functions into another space of functions. Similar to the well-known universal approximation theorem that states neural networks can approximate any continuous function to arbitrary accuracy if no constraint is placed on the width and depth of the hidden layers \cite{hornik1989multilayer}, there is another approximation theorem states that neural networks can accurately approximate any nonlinear operator \cite{chen1995universal}. Such neural networks are called neural operators.

In the meantime, partial differential equations (PDEs) that describe physical phenomena can be viewed as nonlinear operators that map initial conditions (functions defined in Euclidean spaces) to solutions (functions in Euclidean space, with or without time). As a result, a line of research has emerged that utilizes neural operators to approximate the solutions of an entire family of PDEs \cite{bhattacharya2020model, li2020neural,lu2019deeponet}. 
In the context of predicting temperatures during an AM process, simulation and experimental data may be available in different resolutions. Unlike methods based on Convolutional Neural Networks (CNNs) that approximate PDE solutions in discretized Euclidean spaces, which are dependent on the mesh, neural operators can learn solutions with super-resolution. Neural operators trained on a low-resolution mesh can therefore be used to evaluate on a high-resolution mesh. They can thus readily be used to overcome discrepancies between the resolution of simulation and experimental data.
Neural operators are therefore a suitable tool to model and predict the AM process.

The FNO \cite{li2020fourier} is a recently developed type of neural operator that utilizes the fast Fourier transform (FFT) to achieve $nlog(n)$ time complexity, in contrast to the quadratic complexity of other neural operators, like the neural operator proposed in \cite{patel2021physics}. In addition, FNO also features a noise-filtering mechanism brought by spectral analysis. While a brief introduction is provided here, readers are referred to \cite{li2020fourier} for a more detailed review.

Let $D$ be a bounded open subset of $\mathbb{R}^d$, where $d$ is the dimension of the Euclidean space. Let $\mathcal{A}=\mathcal{A}(D; \mathbb{R}^{d_a})$ and $\mathcal{U}=\mathcal{U}(D; \mathbb{R}^{d_u})$ be separable Banach spaces of functions that take values in $\mathbb{R}^{d_a}$ and $\mathbb{R}^{d_u}$, respectively. Typically, $\mathcal{A}$ represents the space of input functions (such as initial conditions), while $\mathcal{U}$ represents the space of output functions (i.e., solutions to PDEs).

Consider a non-linear operator $G^{\dag}: \mathcal{A} \rightarrow \mathcal{U}$ that maps input functions to output functions. Suppose we have a set of $N$ input-output function pairs $\{a_j,u_j\}_{j=1}^N$ observed from $\mathcal{A}$ and $\mathcal{U}$, respectively, where the set of $a$ is selected as an independent and identically distributed sequence from the probability measure $\mu$, and $u = G^{\dag}(a)$. The goal is to approximate $G^{\dag}$ by constructing a parametric non-linear operator $G_{\theta}: \mathcal{A} \rightarrow \mathcal{U}$, where $\theta \in \Theta$ denotes the set of parameters. To this end, we define a cost function $C: \mathcal{U}\times \mathcal{U} \rightarrow \mathbb{R}$ and seek to minimize the expected value of this cost function over the input function space, i.e.,

\begin{equation}
\min_{\theta \in \Theta}\mathbb{E}_{a \sim \mu}[C(G_{\theta}(a),u)].
\end{equation}

It is important to note that in practical applications, we often only have access to the point-wise evaluations of the functions $a$ and $u$ obtained from simulations or experiments. To simplify notation, in the following discussion, we will use $a$ and $u$ to refer to these numerical observations, where $a \in \mathbb{R}^{n\times d_a}$ and $u \in \mathbb{R}^{n\times d_u}$, with $n$ being the number of sample points used to discretize the domain $D$.
As a non-linear operator, $G_{\theta}$ is mesh-independent or super-resolution, which means that it can be used to evaluate functions at positions that are not in the sample points.

Directly parameterizing the non-linear operator $G^\dagger$ can be challenging. A potential solution is to use the idea of contracting neural networks, which approximate any continuous functions by breaking down the calculation into multiple layers and combining a linear transformation with a non-linear activation function in each layer. Similarly, a non-linear operator can be approximated using a series of iterative updates, where each update consists of a global linear operator and a local non-linear activation function. 

The computational efficiency of neural operators is of great importance for their practical applications, as the complexities of global linear operators often lead to significant challenges in achieving efficient computation \cite{li2020fourier}. In recent years, the Fourier neural operator has emerged as a promising approach to address this issue. By implementing convolution as the global linear operator and leveraging the fast Fourier transform (FFT), this neural operator achieves $nlog(n)$ time complexity.

Convolution between two function  $g: D \rightarrow \mathbb{R}^{d_v}$ and $f: D \rightarrow \mathbb{R}^{d_v}$ results in a new function $g \ast f : D \rightarrow \mathbb{R}^{d_v}$, which is defined as  
$$
g \ast f = \int_D g(x-y)f(y)dy.
$$
This mathematical operation can be interpreted as a linear operator that transforms a function $f$ by performing a global integral with another function $g$.

We can parameterize convolution by using a family of parametric kernel functions $k_{\phi}$, where $\phi$ belongs to a given set of parameters $\Theta$. With this parameterization, we can define a global linear operator as follows:
\begin{equation}\label{equ:kernel}
k_{\phi} \ast f(x):= \int_D k_{\phi}(x-y)f(y)dy,
\end{equation}
where $f$ and $k_{\phi}$ are functions defined over a domain $D$. This operator transforms a function $f$ using a global integral with the kernel function $k_{\phi}$.

The FNO calculation process can be summarized as follows. First, the input $a \in \mathcal{A}$ is lifted to a higher dimension using a local linear transformation $P: \mathbb{R}^{d_a} \rightarrow \mathbb{R}^{d_v}$, such that $v_0(x) = P(a(x))$. Next, a series of iterative updates is applied, generating $v_0 \mapsto v_1 ... \mapsto v_T$, where each $v_h$ takes value in $\mathbb{R}^{d_v}$. Finally, the output $u(x) = Q(v_T(x))$ is projected back by a local linear transformation $Q: \mathbb{R}^{d_v} \rightarrow \mathbb{R}^{d_u}$.
The iterative update is defined as
\begin{equation}\label{equ:updates}
    v_{t+1}(x) := \sigma( W v_t(x) + k_{\phi} \ast v_t(x) ), \forall x \in D,
\end{equation}
where $W:\mathbb{R}^{d_v} \rightarrow \mathbb{R}^{d_v} $ is a linear transformation, $\sigma: \mathbb{R} \rightarrow \mathbb{R}$ is a local non-linear activation function.

To enable efficient computation of the convolution operation in Equation \ref{equ:updates}, the Fourier transform is utilized. Specifically, the Fourier transform of a function $f: D \rightarrow \mathbb{R}^{d_v}$ is denoted as $\mathcal{F}(f)$, and its inverse is denoted as $\mathcal{F}^{-1}(f)$. By applying the convolution theorem, the Fourier transform of a convolution of two functions can be expressed as the component-wise product of their Fourier transforms. Thus, we have the following expression for the convolution operation in Fourier space:
$$
    k_{\phi} \ast v_t(x) = \mathcal{F}^{-1}(\mathcal{F}(k_{\phi}) \cdot \mathcal{F}(v_t))(x), \forall x \in D,
$$
where $\cdot$ denotes component-wise multiplication.

Since only finite point-wise evaluations of the function $v_t$ are available, the modes of $v_t$ in Fourier space are finite. In order to filter out noise in frequency, only low frequency modes are retained for the neural operator. Let $\xi_{max}$ denote the number of the modes left after filtering.

Moreover, rather than constructing a family of parametric functions for $k_{\phi}$, a more convenient approach is to directly parameterize $k_{\phi}$ in Fourier space. Consequently, the parameterized linear operator is given by Equation \ref{equ:R}.
\begin{equation}\label{equ:R}
    k_{\phi} \ast v_t(x) = \mathcal{F}^{-1}(R \cdot \mathcal{F}(v_t))(x), \forall x \in D,
\end{equation}
where $R$ is a complex-valued $(\xi_{max} \times d_v \times d_v)-$tensor whose components are the parameters of the linear operator.
When the domain $D$ is discretized uniformly, the FFT algorithm can be applied to efficiently calculate Equation \ref{equ:R} with a complexity of $O(n\log n)$.

\section{Method}\label{sec:method}
\begin{figure}
    \centering
    \includegraphics[width=0.3\linewidth]{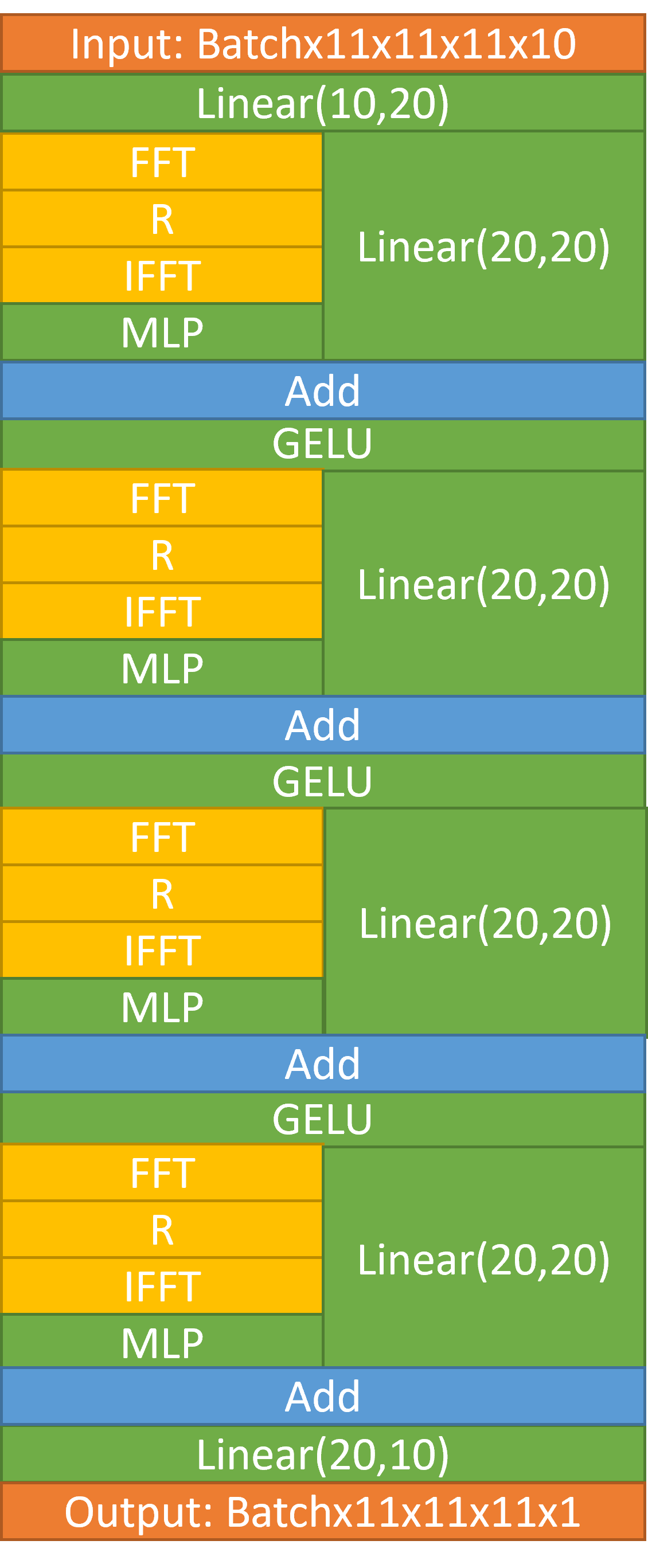}
    \caption{The architecture of the machine learning model. MLP represents the multilayer perceptron. FFT and IFFT represent the fast Fourier Transformation and its inverse. R represents the linear transformation in the complex space. GELU is the nonlinear activation function. }
    \label{fig:ml_arch}
\end{figure}
The architecture of the ML model is shown in Figure \ref{fig:ml_arch}. Note that the ML model does not operate on the whole domain at once, but on the smaller regions, called heat-affected windows, introduced in Section \ref{sec:window}

\subsection{Heat-Affected Windows}\label{sec:window}
Thermal simulations of additive manufacturing processes involve the solution of transient heat transfer partial differential equations (PDEs). Often elements are activated sequentially according to a toolpath prescribed on a pre-defined mesh. For a model meshed with equally-sized elements, the time step $\Delta t$ between successive element activation is then determined by the element size and the tool's moving speed.
 When a new element is activated, the PDEs are solved with the temperature prior to activation and boundary conditions, such as heat influx, convection, and fixed temperature, as input. The output is the temperature after $\Delta t$. We have observed that over a short time period, the evolution of temperature is primarily confined in a small region. Thus, to predict the temperature at a specific position after $\Delta t$, we only need the information of its local neighbor region, not the whole domain. In heat transfer analysis, the thermal diffusivity $\alpha_p$ characterizes a material's rate of heat transfer. For instance, for steel, $\alpha_p$ is approximately $12$ $ mm^2/s$. This means that for $\Delta t=0.1$ $ s$, the area of the neighborhood affecting the temperature of a given position is about $1.2$ $ mm^2$. Therefore, we define the heat-affected neighborhood's characteristic radius as
\begin{equation}
r_c = \sqrt{\alpha_p \Delta t}.
\end{equation}
We can choose a box region around a position whose size is about 10 times $r_c$ to ensure that most of the necessary information is included in the box to predict the temperature for the position. These box regions are called heat-affected windows or windows in this paper.
\begin{figure}
    \centering
    \includegraphics[width=0.7\linewidth]{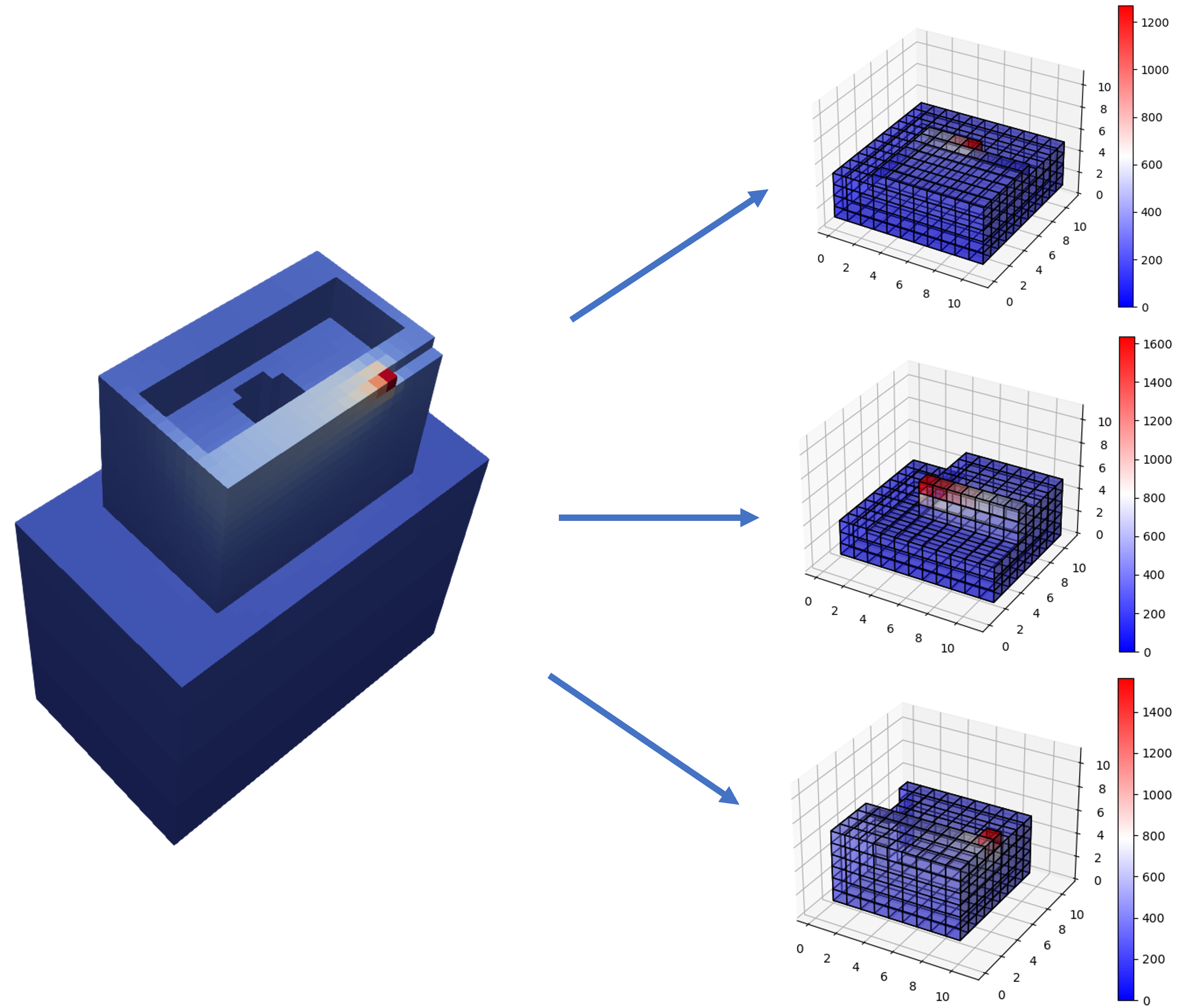}
    \caption{Instead of learning the temperature evolution over the domain, the ML model proposed in this paper works on the local regions cut from the domain, called heat-affected windows.}
    \label{fig:window}
\end{figure}

\begin{figure}
    \centering
    \includegraphics[width=0.6\linewidth]{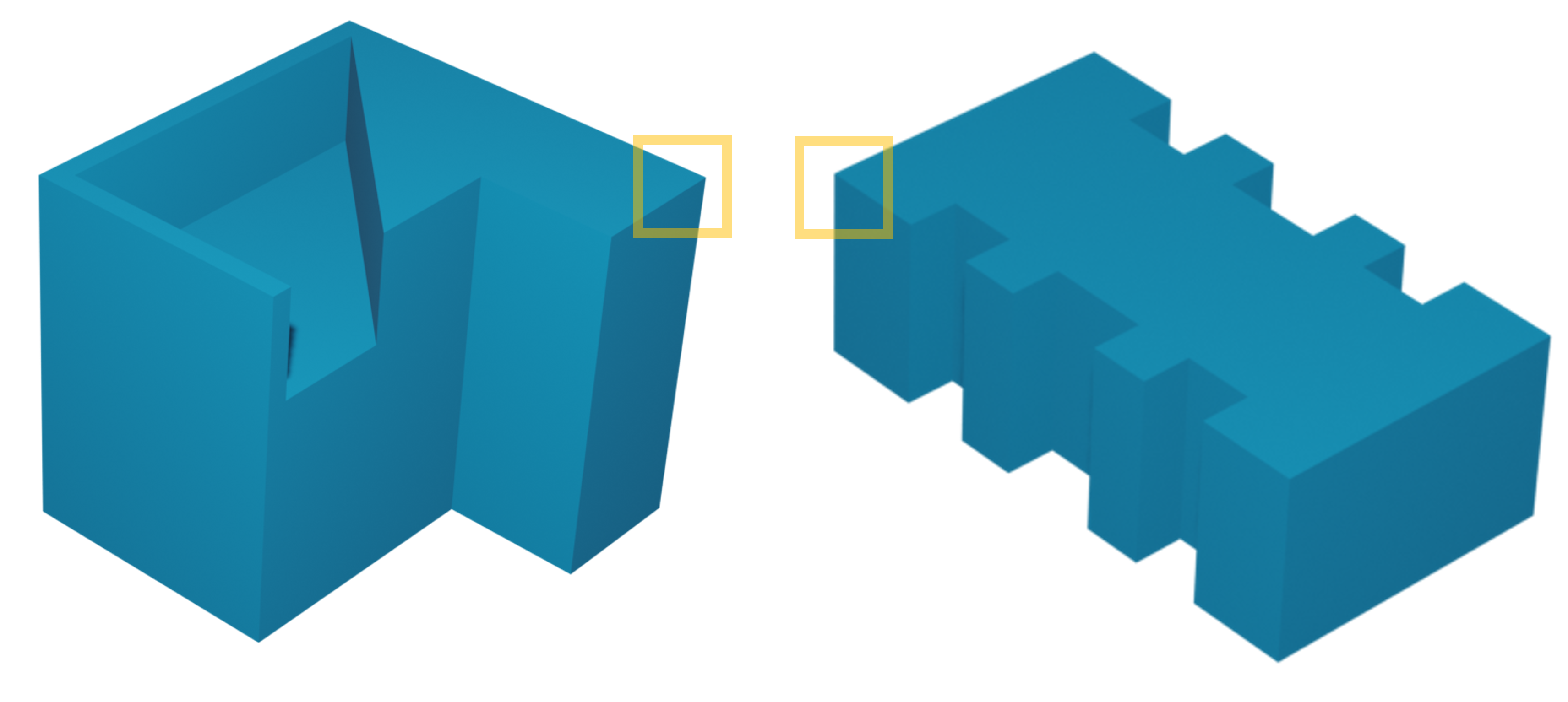}
    \caption{Two different shapes could look similar locally.}
    \label{fig:local_similar}
\end{figure}

In this study, the ML model is designed to operate solely on the heat-affected windows rather than the entire domain. This approach has two significant advantages. Firstly, by reducing the number of parameters, the size of the ML model can be significantly reduced, thereby requiring less training data. Secondly, this technique enhances the generalizability of the model for a wide variety of geometries, even with a limited training dataset. This is because two different geometries may share similar local features despite appearing vastly different. For example, Figure \ref{fig:local_similar} shows two different shapes but their heat-affected windows around the boundary look similar as both of their boundaries consist of plane surfaces.

\subsection{Input Variables}
In addition to the input temperature, the ML model is also provided with other relevant information related to the heat transfer process for each element.

\noindent \textbf{Activation Indicator.} \hspace{0.5 cm} Heat affected windows located near the edges of the part may include void space where no material is present. To handle this case, we introduce a variable $\rho_{act}$ that indicates whether an element is activated. Here, a value of 1 means that there is material, while a value of 0 means that there is a void.

\noindent \textbf{Heat Influx Conditions.} \hspace{0.5 cm} The vector $\mathbf{H}$ contains two pieces of information relevant to the heat influx condition for an element: the power of the input energy and the relative position from the center of the element to the heat influx position.

\noindent \textbf{Boundary Impact Factors.} \hspace{0.5 cm} The simulation considers two types of boundary conditions: convection-radiative and fixed-temperature conditions. The former is applied on the outer surface of the built part, while the latter is applied on the substrate bottom which the part is built on. The substrate bottom temperature is set to room temperature. To describe how each boundary condition affects local temperature evolution, boundary impact factors (BIF) $\mathbf{B}$ are defined for each element. $\mathbf{B}$ has two components corresponding to the two types of boundaries.  Although there could be more complicated ways to define $\mathbf{B}$, a simple distance-based approach is used in this work; namely, the distance between the element center and each type of boundary.

\subsection{Loss Function And Evaluation Metrics}
We use the normalized $L_2$ error as the loss function for training our model, denoted as $NL_2$. Let $u_{pred}$ be the predicted temperature over a window, and $u$ be the ground truth temperature of the window. $u_{{pred}_i}$ and $u_i$ represent the predicted and ground truth temperature of an individual element, respectively. The window is uniformly discretized with $n$ elements. $NL_2$ is defined as 
\begin{equation}
    NL_2 = \sum_{i=1}^n \frac{\sqrt{(u_{{pred}_i} - u_i)^2}}{|u_i|}.
\end{equation}

The mean squared error (MSE) is a commonly used metric for evaluating machine learning models. It is defined as the average of the squared differences between the predicted and true values, calculated as:

\begin{equation}
MSE = \frac{1}{n}\sum_{i=1}^n (u_{{pred}_i} - u_i)^2.
\end{equation}
To assess the performance of ML models that make predictions on varying scales, the normalized root mean squared error (NRMSE) is often used. It is defined as:

\begin{equation}
NRMSE = \sqrt{ \frac{1}{n} \sum_{i=1}^n \left (\frac{u_{{pred}_i} - u_i}{u_i} \right )^2 }.
\end{equation}

Here, the NRMSE takes into account the relative magnitude of the temperature values by normalizing the squared differences with respect to the ground truth temperature.

While MSE and NRMSE can reflect the overall accuracy of the ML model in many cases, they might overestimate the performance of ML models for temperature prediction of AM processes. This is because most of the domain does not experience significant temperature changes except in regions near recent deposition in a time step. Thus, a prediction that fails to capture temperature changes in these regions may still have a good MSE or NRMSE score. To address this issue, we propose to use the $R^2$ metric to evaluate the ML models, which measures the proportion of the variance in the ground truth that is explained by the model's predictions. It is defined as
\begin{equation}
    R^2 =  1 -  \frac{\sum_{i=1}^n (u_{{pred}_i} - u_i)^2}{\sum_{i=1}^n (u_{mean} - u_i)^2},
\end{equation}
where $u_{mean}$ is the average of $u$ over the window.
$R^2$ takes value in $(-\infty,1]$. A negative $R^2$ value indicates that the model's predictions have a higher mean squared error (MSE) than a simple baseline predictor that uses the mean temperature as the prediction. When $R^2$ is close to zero, it suggests that the model's predictions are similar to those of the baseline predictor. Conversely, as $R^2$ approaches 1, the accuracy of the model's predictions improves, indicating a better fit between the model and the observed data.

\section{Dataset Generation}\label{sec:data_generation}
We developed a dataset that consists of the geometric models created by a generative ML model, and a high-fidelity finite element simulation is employed to get the temperature history of all geometric points.

\subsection{Geometric Model Creation}
\begin{figure}[h!]
    \centering
    \includegraphics[width=0.8\linewidth]{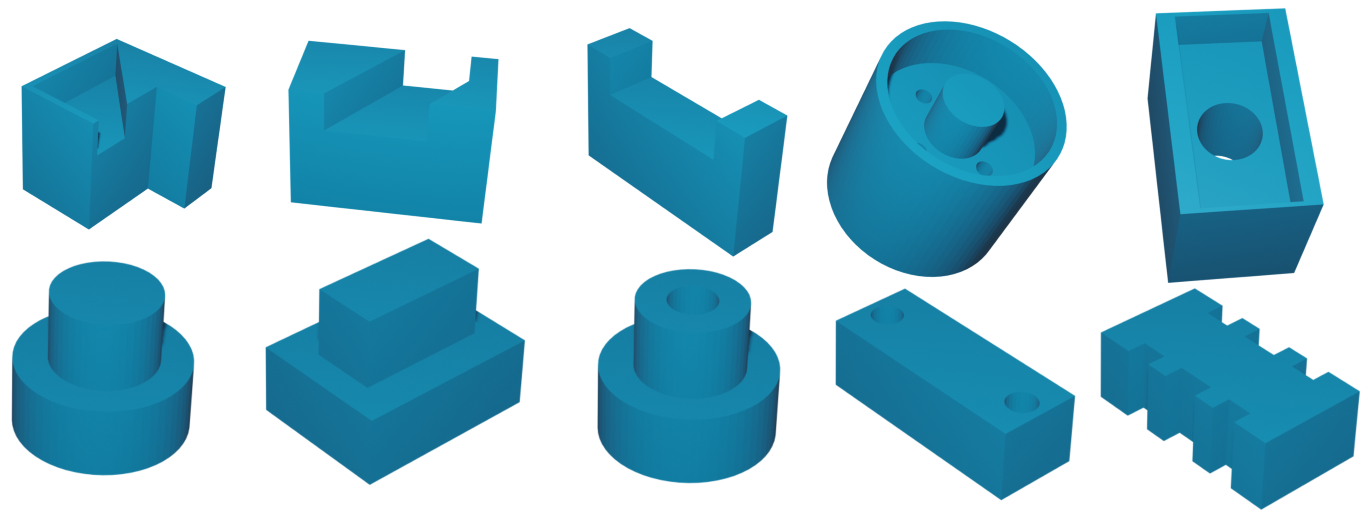}
    \caption{Geometric models created by SkexGen. The bounding boxes of these geometric models are in the dimensions of 40 $mm$ $times$ 40 $mm$ $times$ 40 $mm$.}
    \label{fig:geometric_models}
\end{figure}

SkexGen \cite{xu2022skexgen} is an autoregressive generative model based on transformers that encode topological, geometric, and extrusion variations of CAD model construction sequences into disentangled code books \cite{van2017neural}. With SkexGen, we can randomly generate geometric models with prescribed topological and extrusion features in various geometric details. 
In practice, we can use this CAD generative model to augment a dataset in a limited amount of time by generating synthesized models that share specific similarities with the existing dataset.
Figure \ref{fig:geometric_models} shows the ten models created by SkexGen. Each row is generated with the same extrusion code but with different topologies and geometries. The models in the first row all have carved features in their upper region, while those in the second row are either stacked structures or contain cylindrical holes.

While it is trvial to randomly generate hundreds of geometric models, the scale of the dataset used here is bottle-necked by the intensive computation required by the thermal simulation.

The dimensions of the geometric models are normalized into a similar level before meshing. For all the models, the largest dimensions is set as 40 $mm$. 

\subsection{Thermal Simulation}
Consider a domain $\Omega$ bounded by its boundary $\partial \Omega$, of which $\partial \Omega_\mathrm{H}$ represents the part at which heat is transferred to the surroundings with constant temperature $T_{\infty}$, and $\partial \Omega_\mathrm{D}$ the part at which the temperature is fixed at $T_\mathrm{D}$. The temperature evolution within $\Omega$ is governed by the following set of PDEs:
\begin{equation}\label{equ:pdes}
\begin{aligned}
\rho c_p \dot{T} &= \nabla \cdot (k_p \nabla T), &&\forall \mathbf{x} \in \partial \Omega  \\
-\mathbf{n} \cdot k_p \nabla T &= h_\mathrm{c}(T-T_{\infty}), &&\forall \mathbf{x} \in \partial \Omega_\mathrm{H} \\
T &= T_\mathrm{D}, &&\forall \mathbf{x} \in \partial \Omega_\mathrm{D}.
\end{aligned}
\end{equation}
Here, $\rho$, $c_\mathrm{p}$ and $k_\mathrm{p}$ are the temperature-dependent density, specific heat capacity and conductivity of the material, respectively. The vector $\mathbf{n}$ is the unit outward normal of the boundary at coordinate $\mathbf{x}$. In Equation \ref{equ:pdes}, $h_\mathrm{c}$ is a temperature-dependent heat transfer coefficient that accounts for free convection with  convection coefficient $h_\infty=15 W/\left(m^2K\right)$ and radiation with emissivity $\epsilon=0.35$ as
\begin{equation}\label{equ:convrad}
    h_\mathrm{c}(T) = \epsilon\sigma_\mathrm{b} \left( T^3+T^2T_\infty +T T_\infty^2 +T_\infty^3  \right),
\end{equation}
where $\sigma_\mathrm{b}$ is the Stefan-Boltzmann constant.

We utilized the thermal simulation algorithm developed in \cite{nijhuis2021efficient} to efficiently and accurately solve the partial differential equations described in Equation \ref{equ:pdes} for the DED process. This algorithm uses the discontinuous Galerkin finite element method (DGFEM) to spatially discretize the problem and the explicit forward Euler timestepping scheme to advance the solution in time. The algorithm activates elements based on the predefined toolpath. Newly-deposited elements are initialised at elevated temperature, after which they are allowed to cool according to Equation \ref{equ:pdes}. To ensure that the high process heat input is captured correctly, newly-deposited elements are assigned an enhanced heat capacity prior to their solidification.

Our simulations utilized S355 structural steel as the material, with material properties as given in \cite{nijhuis2021efficient}. The activation temperature of newly-added elements was set to $1750^\circ C$ and the enhanced specific heat capacity to $4537.9 J/(kgK)$. The temperature of the substrate’s bottom face is kept fixed at $T_\infty =25^\circ C$. On all other faces, convection and radiation to the surrounding air at $T_\infty$ is modelled using Equation \ref{equ:convrad}. We set the tool moving speed to $5 mm/s$. All geometric models were discretized with a resolution of $20\times20\times20$, with an element size of $2 mm$.

\subsection{Data Preprocess and Training Setting}
\begin{figure}
    \centering
    \includegraphics[width=0.8\linewidth]{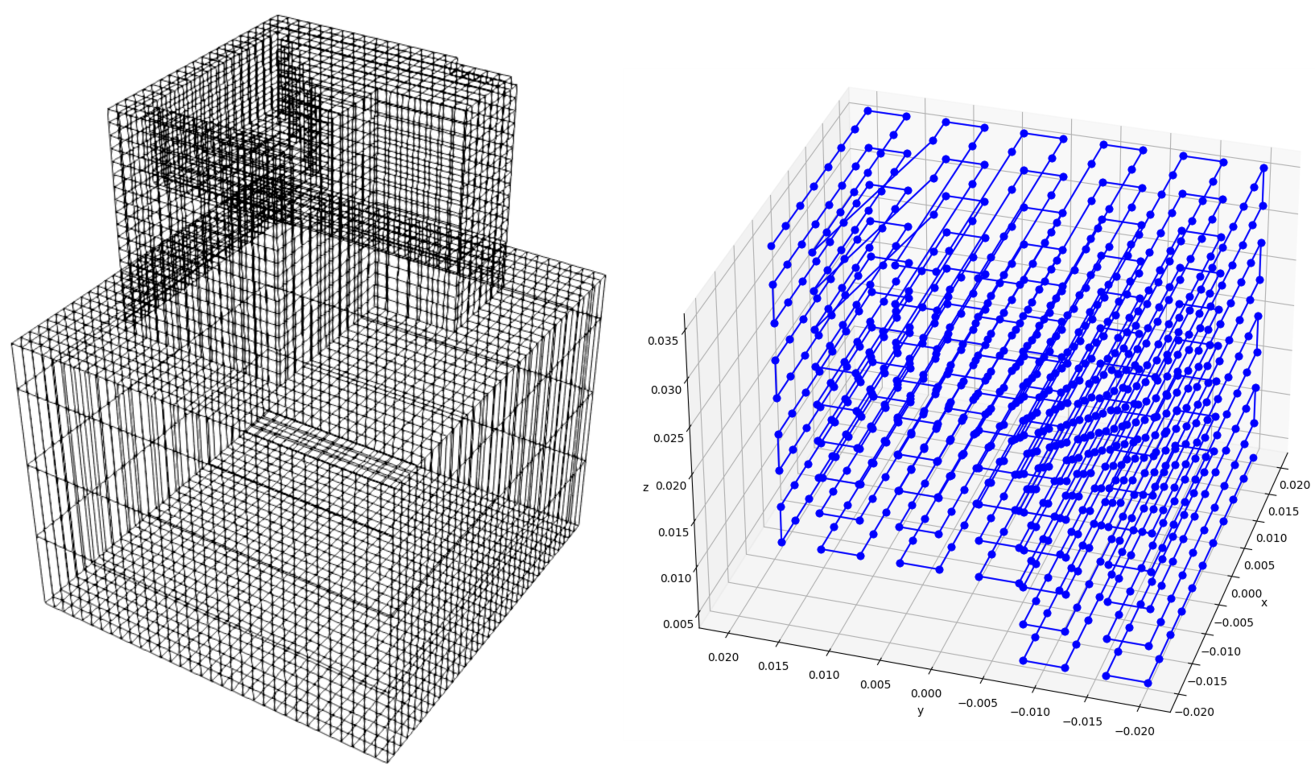}
    \caption{The mesh (with substrate) and the ``zigzag'' toolpath automatically generated by the algorithm.}
    \label{fig:mesh_toolpath_example}
\end{figure}

An Efficient software for data preprocessing of thermal simulation have been developed by the authors to generate hexahedron meshes from geometric models created by SkexGen. It can add a coarse mesh for the substrate. Additionally, a toolpath that specifies the locations of the deposition tool in a sequence of time steps can be constructed based on the process parameters. The code is publicly available on GitHub \footnote{\href{https://github.com/Jiangce2017/hammer_chuizi.git}{https://github.com/Jiangce2017/hammer\_chuizi.git}}. An example of the mesh and toolpath generated by the code is shown in Figure \ref{fig:mesh_toolpath_example}.

The simulation generates the temperatures of all activated elements at each time step. Let $E = {e_0, e_1, ..., e_m}$ be the set of elements ordered by their activation sequence, where $e_i$ is the element deposited at time step $t_i$. The temperature at time step $t_i$ serves as the input to predict the temperature at time step $t_{i+1}$, which is calculated by the numerical simulation and serves as the ground truth for the prediction. To focus the ML model on the temperature evolution in the HAZ near the most recent deposition, heat-affected windows with dimensions $11\times 11\times 11$ are constructed around the deposited elements. Specifically, windows around the elements ${e_{i-9}, e_{i-8}, ..., e_i}$ are initially selected to train the ML model. Subsequently, more windows can be included to allow the ML model to predict the temperature of the entire domain. By doing so, we can let the ML model prioritize on the HAZ of the recent deposition. 
\begin{table}[]
    \centering
    \begin{tabular}{c|c|c}
       Index & Geometry & Number of Windows  \\
        \hline
        1 & \includegraphics[width=0.1\linewidth]{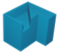} &  9819 \\
        2 & \includegraphics[width=0.1\linewidth]{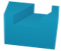} & 9804 \\
        3 & \includegraphics[width=0.1\linewidth]{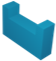}& 9784 \\
        4 & \includegraphics[width=0.1\linewidth]{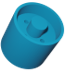}& 9513 \\
        5 & \includegraphics[width=0.1\linewidth]{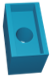}& 9707 \\
        6 & \includegraphics[width=0.1\linewidth]{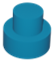}& 9921 \\
        7 & \includegraphics[width=0.1\linewidth]{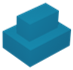}& 9977 \\
        8 & \includegraphics[width=0.1\linewidth]{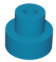}& 9588 \\
        9 & \includegraphics[width=0.1\linewidth]{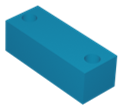}& 9825 \\
        10 & \includegraphics[width=0.1\linewidth]{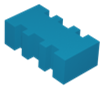}& 9725\\
    \end{tabular}
    \caption{The number of the windows selected from the thermal simulation of each geometric model.}
    \label{tab:wind_number}
\end{table}
Table \ref{tab:wind_number} provides the number of windows for each geometric model. To maximize the use of our limited dataset, we performed a procedure similar to k-fold cross-validation to evaluate the generalizability of the ML model to unseen geometries during training. We conducted 10 rounds of training and validation, with each round using a different geometry for validation and the remaining nine for training and testing. The windows from the nine geometries were mixed and randomly divided into two sets: 90\% for training and 10\% for testing. We trained the model for 50 epochs using Adam as optimizer with a learning rate of $1\times 10^{-3}$ and weight decay of $1\times 10^{-4}$. After training, we used the ML model to predict the temperatures of the validation geometry. 

\section{Results and Discussion}\label{sec:results}
Table \ref{tab:train_results} displays the training and test metrics after 50 training epochs of each cross-validation round. The $MSE$, $NL_2$, and $R^2$ values are the average quantities across all windows in the training, test, or validation dataset. The high degree of consistency between the training and testing metrics indicates that the ML model is capable of accurately capturing temperature evolution without overfitting. It should be noted that the ground truth and prediction temperatures used for evaluation are their original values without normalization, and the temperature unit is Celsius. The relatively large variance of $MSE$ is likely due to the variance of temperature responses in the AM process across different geometries.

\begin{figure*}
    \centering
    \includegraphics[width=0.8\linewidth]{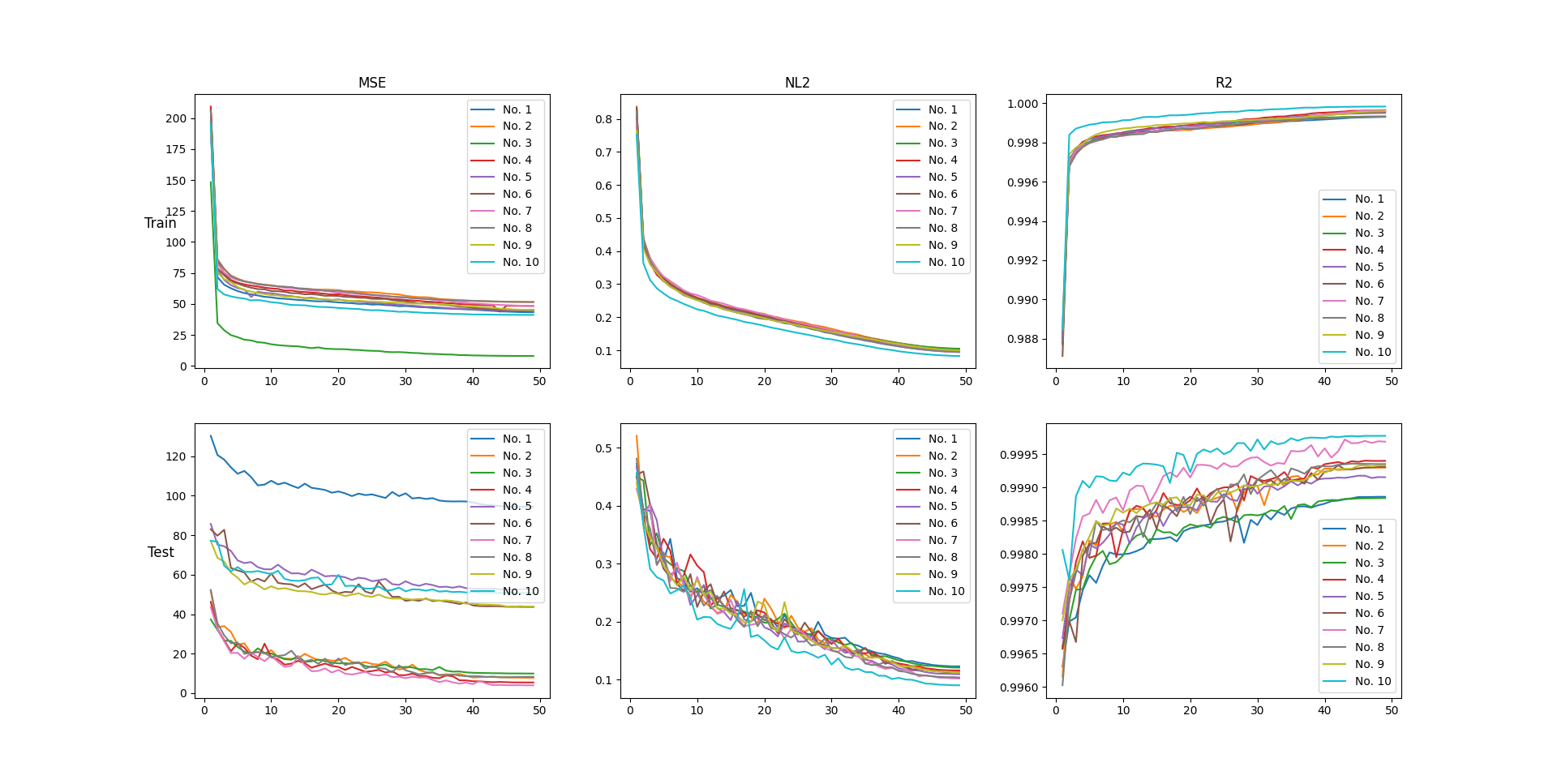}
    \caption{The history of the training and test evaluation metrics during the 50 epochs training.}
    \label{fig:train_history}
\end{figure*}

Figure \ref{fig:train_history} shows the histories of these metrics during the training process. $MSE$ and $NL_2$ decrease rapidly in all rounds, and $R^2$ quickly approaches 1, indicating that the optimizer converges successfully. $MSE$ provides a straightforward way to describe the average error square between the prediction and ground truth. For example, the test $MSE$ of No.1 round converges to about 100, indicating that the average absolute difference between prediction and ground truth is approximately $10^{\circ} C$ which is an acceptable error as the highest temperature could achieve $1500^{\circ} C$ in the simulation of AM process.
$NL_2$ measures a comparative error that is divided by the value of the ground truth. This is why the cross-validation rounds approach similar values of $NL_2$ at the end. However, interpreting the ML model's performance in capturing the dramatic temperature evolution from $NL_2$ and $MSE$ is challenging if most windows are far from the recent deposition and do not have significant temperature changes.
$R^2$ reveals the relative accuracy of the prediction compared with using the mean of ground truth as the prediction. In other words, it measures the proportion of the variance in the ground truth that the prediction explains. The fact that the test $R^2$ in all rounds is above $0.99$ suggests that the model can capture the extreme local temperature variance.

\begin{figure*}
    \centering
    \includegraphics[width=0.75\linewidth]{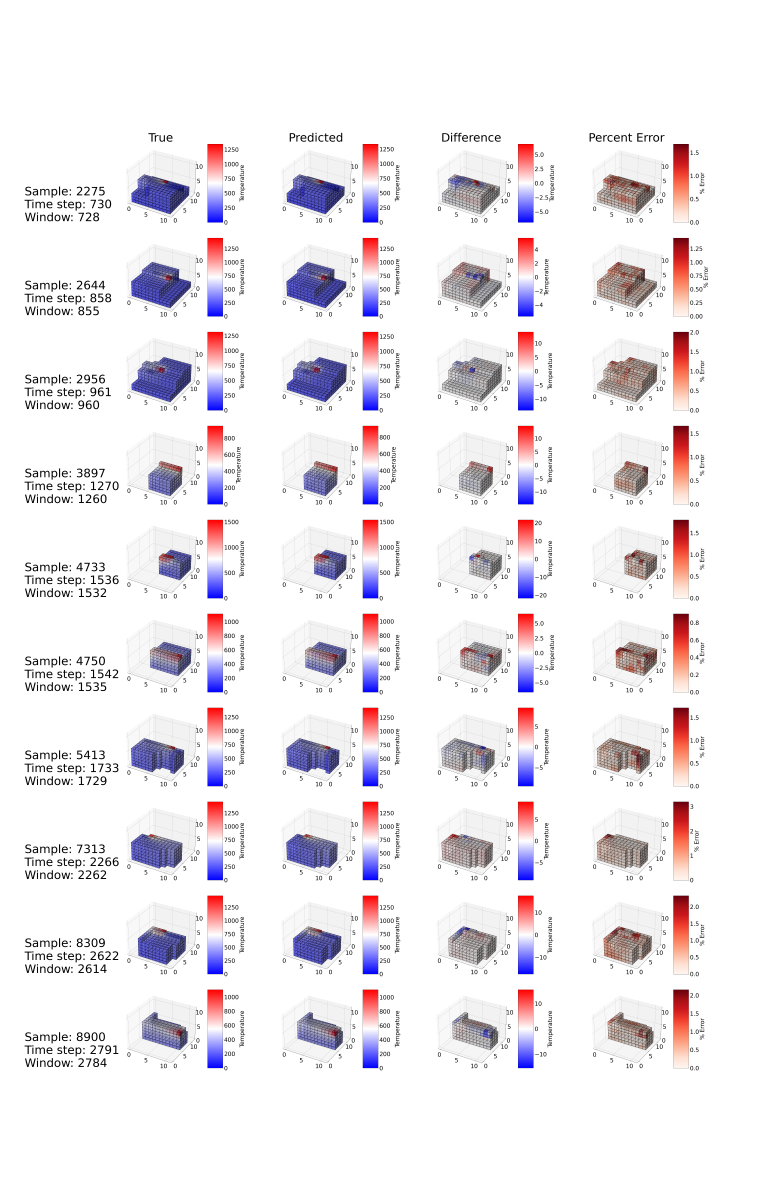}
    \caption{The visualization of the prediction performance of the ML model trained validated on geometric model $5$. There are 10 randomly selected windows in different timesteps. From the left to the right, each column shows the ground truth of the temperature, the prediction of the temperature, the different between  the ground truth and the prediction,and the percent error, respectively. The temperature unit is Celsius.}
    \label{fig:50_epoch}
\end{figure*}

 Table \ref{tab:valid_table} lists the performance of the ML model on the validation geometric models, which are unseen in the training process. 
The ML model shows good generalizability in 7 out of 10 geometries, but fails significantly in 3 of the 10. Figure \ref{fig:50_epoch} visualizes the prediction results of the validation on geometric model $5$. 10 windows are randomly selected for each row. At each row, the ground truth, the predicted temperature, the difference between the prediction and the temperature, and the error percent distribution over voxels of the window are shown.  As we can see, the model can predict the temperature precisely for this validation geometric model, with errors ranging within 3\% for most of the windows. The largest errors tend to appear near the recent deposition, with temperatures there being slightly overestimated by the ML model.

The ML model appears to fail completely at predicting the temperature for geometric models $6$, $7$, and $8$, despite performing well in learning their local temperature evolution when included in the training process. To investigate how prediction errors are distributed over the windows, we analyzed the $R^2$ of the 10 windows with the worst predictions (lowest $R^2$) in each cross-validation round, as shown in Table \ref{tab:10_worst_window}. Among approximately 10 thousand windows, only a few poorly predicted windows can adversely affect the average prediction.

\begin{figure*}
    \centering
    \includegraphics[width=0.75\linewidth]{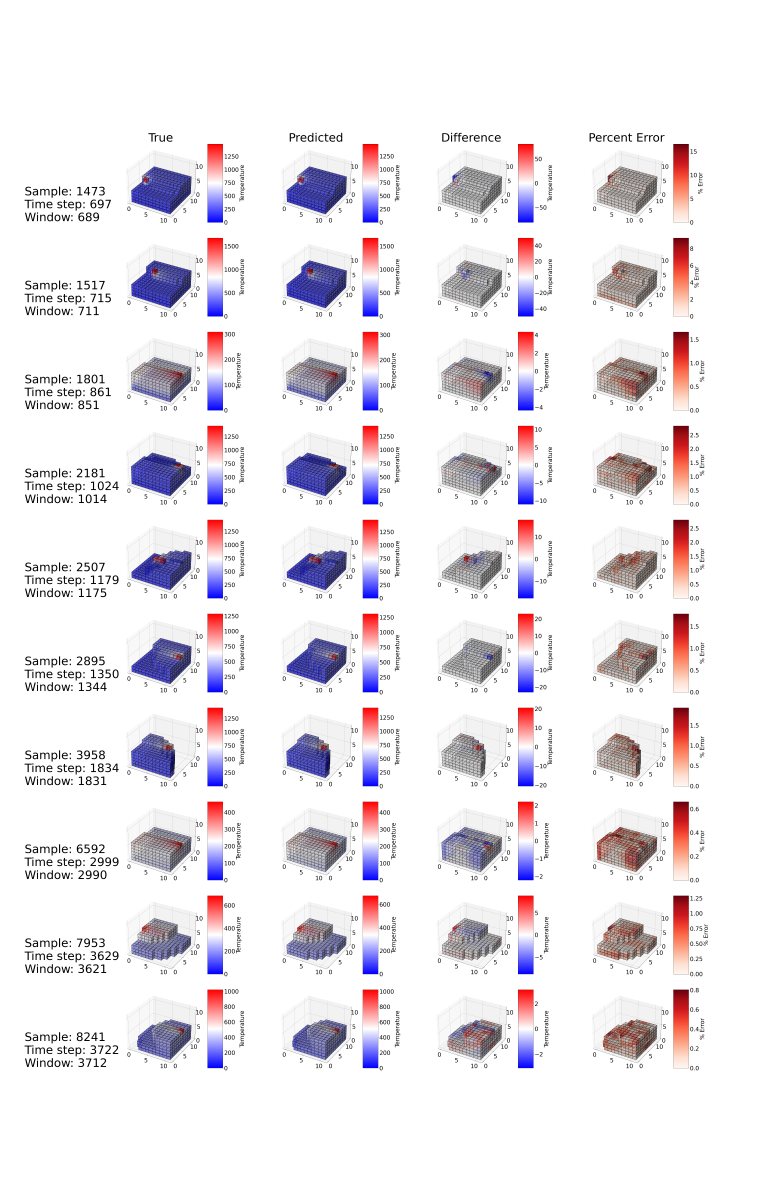}
    \caption{The visualization of the prediction performance of the ML model trained validated on geometric model $6$.}
    \label{fig:validated_on_6}
\end{figure*}

Figure \ref{fig:validated_on_6} depicts the predictions of 10 randomly selected windows from the cross-validation rounds on geometric model $6$, which show a similar level of accuracy as those in Figure \ref{fig:50_epoch}. Therefore, despite the three failed cross-validation rounds, most of the windows have reliable predictions. These observations align with the findings reported by \cite{to2023modified}, which suggests that most points in the parts built by the AM process experience repetitive and similar temperature histories.


This also suggests that the three geometries possess unique local features that are not present in the other geometries used in training. Thus, the current dataset may be insufficient for covering the range of common geometric models used in practice. Additionally, the large errors in these cases indicate that the ML model may overfit to the training geometries if certain representative geometries are not included. To overcome these limitations, we suggest building a larger dataset that includes comprehensive geometric features of parts built by the AM process.

\begin{table*}[]
    \centering
    \begin{tabular}{c|c|c|c|c|c|c|c}
      Index & Geometry   & Train $MSE$ & Train $NL_2$ & Train $R^2$ & Test $MSE$ & Test $NL_2$ & Test $R^2$  \\
    \hline
      1 & \includegraphics[width=0.05\linewidth]{images/hollow_1.png}  &  43.2577 & 0.1024 & 0.9993 & 94.3523 & 0.1226 & 	0.9989 \\
      2 & \includegraphics[width=0.05\linewidth]{images/hollow_2.png} & 51.4106& 0.1038 & 0.9993 & 7.8131 &	0.1128 &	0.9993 \\
      3 & \includegraphics[width=0.05\linewidth]{images/hollow_3.png}& 7.9197 & 	0.1047 &0.9993 & 9.9346 & 	0.1203 & 0.9988 \\
      4 & \includegraphics[width=0.05\linewidth]{images/hollow_4.png}& 48.4153 & 0.0991 & 0.9996 & 5.4243 &  0.1158 & 0.9993 \\
      5 & \includegraphics[width=0.05\linewidth]{images/hollow_5.png}& 44.5454 & 0.0944 & 	0.9995 & 52.5766 & 0.1089 & 0.9991 \\
      6 & \includegraphics[width=0.05\linewidth]{images/townhouse_2.png}& 45.1410 & 0.0978 & 0.9995 & 43.6450 & 0.1119 & 0.9993 \\
      7 & \includegraphics[width=0.05\linewidth]{images/townhouse_3.png}& 48.4310 & 0.0947 & 	0.9996 & 4.0064 & 0.1016 & 	0.9997 \\
      8 & \includegraphics[width=0.05\linewidth]{images/townhouse_5.png}& 51.5494 & 0.0954 & 0.9993 & 8.2268 & 0.1033 & 0.9993 \\
      9 & \includegraphics[width=0.05\linewidth]{images/townhouse_6.png}& 45.1011 & 0.0996 & 0.9996 & 43.8328 & 0.1124 & 0.9993 \\
      10 & \includegraphics[width=0.05\linewidth]{images/townhouse_7.png}& 41.1014 & 0.0826 & 0.9998 & 50.6812 & 0.0898 & 0.9998 \\
    \end{tabular}
    \caption{The training and test metrics at after 50 epochs of each cross-validation round}
    \label{tab:train_results}
\end{table*}

\begin{table}[]
    \centering
    \begin{tabular}{c|c|c|c}
     Index & Geometry    & $MSE$ & $R^2$  \\
      \hline
     1 & \includegraphics[width=0.1\linewidth]{images/hollow_1.png}    &  7.362310 & 0.999586 \\
     2 & \includegraphics[width=0.1\linewidth]{images/hollow_2.png}    &  0.813440 & 0.999953 \\
     3 & \includegraphics[width=0.1\linewidth]{images/hollow_3.png}    &  131.311879 & 0.997899 \\
     4 & \includegraphics[width=0.1\linewidth]{images/hollow_4.png}    &  2.830105 & 0.999730 \\
     5 & \includegraphics[width=0.1\linewidth]{images/hollow_5.png}    &  1.963513 & 0.999787 \\
     6 & \includegraphics[width=0.1\linewidth]{images/townhouse_2.png}    &  \textcolor{red}{$4.438254\times10^{11}$} & \textcolor{red}{$-2.433204\times 10^7$} \\
     7 & \includegraphics[width=0.1\linewidth]{images/townhouse_3.png}    &  \textcolor{red}{$1.027052\times10^{11}$} & \textcolor{red}{$-7.695437\times 10^6$} \\
     8 & \includegraphics[width=0.1\linewidth]{images/townhouse_5.png}    &  \textcolor{red}{$8.096904\times10^{10}$} & \textcolor{red}{$-4.151847\times10^6$} \\
     9 & \includegraphics[width=0.1\linewidth]{images/townhouse_6.png}    &  49.322650 & 0.995543 \\
     10 & \includegraphics[width=0.1\linewidth]{images/townhouse_7.png}    &  271.619524 & 0.974972 \\
    \end{tabular}
    \caption{The $MSE$ and $R^2$ of each cross validation fold. The geometry shown in each row is the sample held out during training process for validation.}
    \label{tab:valid_table}
\end{table}

\begin{table}[]
    \centering
    \begin{tabular}{c|c|c|c|c|c|c|c}
        Index & 1st & 2nd & 3th & 4th & 5th & 6th & 7th  \\
        \hline
        1 & 0.6698 &  0.6761 & 0.7180 & 0.8004 & 0.8875 & 0.8897 &  0.9064  \\
        2 & 0.9981 &  0.9983 & 0.9987 &  0.9988 & 0.9989 &  0.9989 & 0.9989 \\
        3 & 0.1713 & 0.1840 & 0.6392 & 0.6792 & 0.7008 &  0.7270 &  0.7440 \\
        4 & 0.7132 &  0.7459 & 0.8119 &  0.9417 &  0.9428 & 0.9630 & 0.9722 \\
        5 & 0.7993 &  0.8282 & 0.9115 & 0.9178 & 0.9285 & 0.9491 & 0.9501 \\
        6 & \textcolor{red}{$-8\times 10^{10}$} & \textcolor{red}{$-8\times 10^{10}$} & \textcolor{red}{$-8\times 10^{10}$} 
           & \textcolor{red}{$-5\times 10^{10}$} & 
        \textcolor{red}{$-4\times 10^{10}$} & \textcolor{red}{$-2\times 10^{10}$} & \textcolor{red}{$-9\times 10^9$} \\ 
        7 & \textcolor{red}{$-4 \times 10^{10}$} & \textcolor{red}{$-3\times10^{10}$} &  0.8019 &  0.9966 & 0.9968 & 0.9976 &  0.9979\\
        8 & \textcolor{red}{$-4\times 10^{10}$} &  0.8593 &  0.8671 &  0.8702 & 0.8718 &  0.8725 &  0.8868 \\
        9 & 0.3713 & 0.3943 &  0.4016 &  0.4184 &  0.4310 & 0.4460 & 0.4508 \\
        10 & 0.1106 & 0.1170 & 0.1281 & 0.1480 & 0.1566 & 0.1949 &  0.2295\\
    \end{tabular}
    \caption{The $R^2$ of the 7 worst window temperature predicted in the 10 cross-validation rounds.}
    \label{tab:10_worst_window}
\end{table}

\section{Conclusion}\label{sec:conclude}
This paper presented a data-driven model that uses Fourier Neural Operator to capture the local temperature evolution during the additive manufacturing process. To prepare the training data, we employed an automatic pipeline that uses an autoregressive generative model, SkexGen, to randomly generate a diverse set of CAD models with variations in topological, geometric, and extrusion features. The toolpath and hexahedral mesh for finite element method (FEM) analysis were then generated using a code we developed. The resulting data was used to run high-fidelity, physics-based simulations using DGFEM. The simulations produced ground truth data which was then used to train the ML model.

Our experiments demonstrate that the proposed ML model can accurately capture local temperature changes, irrespective of the geometry. The $R^2$ metric reveals that the model can precisely predict the large variance of temperature distributions in heat-affected zones near recent depositions. However, our experiments also highlight certain limitations. Cross-validation experiments reveal that the model may fail on geometric models that are significantly different from those used in the training dataset, indicating overfitting to the training data. This may be due, in part, to the relatively small size of the current geometric model dataset. Future work will focus on creating a larger dataset to mitigate this issue. Additionally, our method currently uses only one type of toolpath and consistent tool moving speed. Inclusion of various toolpaths, power, and tool speeds is critical to improve model performance, and will be considered in future work. The architecture of the ML model would need to be modified if more external parameters are required as the dimensions of the input change. However, the total model does not need to be re-trained from scratch. With transfer learning methods \cite{zhuang2020comprehensive}, the layers that connect with the input layer need to be replaced and be fine-tuned with other layers on the new data. In addition, to enhance the fidelity of the ML model and accelerate the learning process, the physics-informed neural networks \cite{raissi2017physics,zhu2021machine} and model order reduction methods \cite{wang2023model, wang2022surrogate} could be included into our framework. 

\section*{Acknowledgements}
This material is based upon work supported by the National Science Foundation through Grant No. CMMI-1825535 and by Carnegie Mellon University’s Manufacturing Futures Institute through the MFI Postdoctoral Fellowship Program. Any opinions, findings, conclusions, or recommendations expressed in this paper are those of the authors and do not necessarily reflect the views of the sponsors.

\clearpage
\bibliographystyle{unsrt}

\bibliography{ref} 
\end{document}